\documentclass[lettersize,journal]{IEEEtran}
\usepackage{amsmath,amsfonts}
\usepackage{algorithmic}
\usepackage{algorithm}
\usepackage{array}
\usepackage[caption=false,font=normalsize,labelfont=sf,textfont=sf]{subfig}
\usepackage{textcomp}
\usepackage{stfloats}
\usepackage{url}
\usepackage{verbatim}
\usepackage{graphicx}
\usepackage{cite}
\usepackage{mathabx}
\usepackage[pagebackref,breaklinks,colorlinks]{hyperref}
\usepackage{tikz}
\usetikzlibrary{shapes,arrows}
\usetikzlibrary{positioning}

\usetikzlibrary{chains, decorations.markings, shadows, shapes.arrows,shapes, fit}
\usepackage{circuitikz}
\usepackage{multirow}
\usepackage{overpic}
\usepackage{fontenc}
\usepackage{amssymb}

\newcommand{\etal}{\textit{et al}.,}
\newcommand{\ie}{\textit{i}.\textit{e}., }
\newcommand{\eg}{\textit{e}.\textit{g}.\ }

\makeatletter
\pgfkeys{/pgf/.cd,
  parallelepiped offset x/.initial=2mm,
  parallelepiped offset y/.initial=2mm
}
\pgfdeclareshape{parallelepiped}
{
  \inheritsavedanchors[from=rectangle] 
  \inheritanchorborder[from=rectangle]
  \inheritanchor[from=rectangle]{north}
  \inheritanchor[from=rectangle]{north west}
  \inheritanchor[from=rectangle]{north east}
  \inheritanchor[from=rectangle]{center}
  \inheritanchor[from=rectangle]{west}
  \inheritanchor[from=rectangle]{east}
  \inheritanchor[from=rectangle]{mid}
  \inheritanchor[from=rectangle]{mid west}
  \inheritanchor[from=rectangle]{mid east}
  \inheritanchor[from=rectangle]{base}
  \inheritanchor[from=rectangle]{base west}
  \inheritanchor[from=rectangle]{base east}
  \inheritanchor[from=rectangle]{south}
  \inheritanchor[from=rectangle]{south west}
  \inheritanchor[from=rectangle]{south east}
  \backgroundpath{
    \southwest \pgf@xa=\pgf@x \pgf@ya=\pgf@y
    \northeast \pgf@xb=\pgf@x \pgf@yb=\pgf@y
    \pgfmathsetlength\pgfutil@tempdima{\pgfkeysvalueof{/pgf/parallelepiped offset x}}
    \pgfmathsetlength\pgfutil@tempdimb{\pgfkeysvalueof{/pgf/parallelepiped offset y}}
    \def\ppd@offset{\pgfpoint{\pgfutil@tempdima}{\pgfutil@tempdimb}}
    \pgfpathmoveto{\pgfqpoint{\pgf@xa}{\pgf@ya}}
    \pgfpathlineto{\pgfqpoint{\pgf@xb}{\pgf@ya}}
    \pgfpathlineto{\pgfqpoint{\pgf@xb}{\pgf@yb}}
    \pgfpathlineto{\pgfqpoint{\pgf@xa}{\pgf@yb}}
    \pgfpathclose
    \pgfpathmoveto{\pgfqpoint{\pgf@xb}{\pgf@ya}}
    \pgfpathlineto{\pgfpointadd{\pgfpoint{\pgf@xb}{\pgf@ya}}{\ppd@offset}}
    \pgfpathlineto{\pgfpointadd{\pgfpoint{\pgf@xb}{\pgf@yb}}{\ppd@offset}}
    \pgfpathlineto{\pgfpointadd{\pgfpoint{\pgf@xa}{\pgf@yb}}{\ppd@offset}}
    \pgfpathlineto{\pgfqpoint{\pgf@xa}{\pgf@yb}}
    \pgfpathmoveto{\pgfqpoint{\pgf@xb}{\pgf@yb}}
    \pgfpathlineto{\pgfpointadd{\pgfpoint{\pgf@xb}{\pgf@yb}}{\ppd@offset}}
  }
}
\makeatother

\begin{document}

\title{A Volumetric Saliency Guided Image Summarization for RGB-D Indoor Scene Classification}

\author{\IEEEauthorblockN{Preeti Meena, ~\IEEEmembership{Student Member,~IEEE,}
Himanshu Kumar, ~\IEEEmembership{Member,~IEEE,}
Sandeep Yadav, ~\IEEEmembership{Member,~IEEE,} }
\thanks{Manuscript received September 10, 2023; revised September 19, 2023.

Preeti Meena, Himanshu Kumar, and Sandeep Yadav are with Discipline of Electrical Engineering, Indian Institute of Technology Jodhpur, Jodhpur-342037, India {\tt\small meena.52@iitj.ac.in}, {\tt\small hkumar@iitj.ac.in}, {\tt\small sy@iitj.ac.in}}}

\markboth{Journal of \LaTeX\ Class Files,~Vol.~14, No.~8, August~2021}%
{Shell \MakeLowercase{\textit{et al.}}: A Sample Article Using IEEEtran.cls for IEEE Journals}


\maketitle
\begin{abstract}
Image summary, an abridged version of the original visual content, can be used to represent the scene. Thus, tasks such as scene classification, identification, indexing, etc., can be performed efficiently using the unique summary. Saliency is the most commonly used technique for generating the relevant image summary. However, the definition of saliency is subjective in nature and depends upon the application. Existing saliency detection methods using RGB-D data mainly focus on color, texture, and depth features. Consequently, the generated summary contains either foreground objects or non-stationary objects. However, applications such as scene identification require stationary characteristics of the scene, unlike state-of-the-art methods. This paper proposes a novel volumetric saliency-guided framework for indoor scene classification. The results highlight the efficacy of the proposed method. 

\end{abstract}

\begin{IEEEkeywords}
Image summarization, RGB-D image, Volumetric saliency, Saliency. 
\end{IEEEkeywords}

\section{Introduction}
\label{sec:intro}
\IEEEPARstart{A}{utomatic} computer vision techniques, along with storing, indexing, and searching, can be efficiently performed with contextual representative information. Estimating this representative information of a given image is referred to as Image Summarization \cite{meena2023review}. The image summarization aims to create a more compact yet informative version of the original content for a given application.

RGB-D indoor image summarization is a challenging problem and
can be accomplished by extracting salient objects from the collection of objects in a scene. 
Salient object extraction algorithms can be top-down or bottom-up \cite{wang2019iterative}. The top-down algorithm such as \cite{wang2019iterative} involves learning of visual knowledge, including high-level context-dependent features like face, human, vehicle, text, etc. In contrast, the bottom-up algorithm \cite{wang2019iterative} relies on features of image regions (\eg intensity, color, texture, size, contrast, brightness, position, motion, shape, and orientation) relative to their surroundings.


 Different methods identify salient regions of an image differently based on the employed algorithm and the features used for saliency computation. As a result, the saliency guided summaries, generated using different methods, result in salient regions of different sizes, shapes, and locations. Hence, the selection of an algorithm for saliency computation is closely related to the desired application. Also, perceptual aspects of the human visual system play a key role in determining the salient regions in the image.

Many previous works, including top-down and bottom-up saliency detection models, have achieved remarkable progress.
But, most of the state-of-the-art methods \cite{aytekin2015visual}, \cite{park2022saliency} primarily utilize color, texture, and depth features in the perceptual framework. Consequently, the generated image summary usually contains mostly foreground objects and lacks scene-distinguishing background objects. Thus, such a summary does not perform well in applications requiring scene-related information such as indoor scene classification. In addition, several works utilized volume information \cite{lara2018computational, hou2022multi} for saliency detection to identify the scene-related information. However, these methods completely overlook the other important visual information of the scene and hence may not be effective under different environmental conditions. Thus, a saliency detection approach is required to jointly capture the scene-defining regions present in the foreground and background.

In order to extricate the scene-defining objects as salient regions regardless of their position in RGB-D images, we introduce a new approach that combines bottom-up and top-down mechanisms and produces an image summary using the extracted salient regions. Unlike state-of-the-art methods, this method is based on both volumetric and visual features to compute the saliency.  For effective summarization, apart from volumetric information, regions or segment contours are used as another supervision information to guide the summarization framework. We adopt a parametric model-based segmentation approach that utilizes joint color-spatial-directional-texture information. Thus, the proposed method results in a methodology suitable for detecting scene-characterizing regions present in the foreground as well as in the background of an image. The main contributions of the paper are:
\begin{itemize}
    \item This paper introduces a volumetric saliency guided summarization framework for indoor RGB-D scene classification. The salient objects in the $3$D scene are found using visual saliency guided by the volumetric saliency. The obtained results highlight the efficacy of the proposed method for the scene classification task.
    
    \item The paper also introduces a two-stage probabilistic fusion framework of geometric (\eg spatial and direction) and visual features (\eg color and texture) to delineate the boundary of the objects within RGB-D images. This additional information assists in the saliency scores computation for volumetric saliency guided summarization.

\end{itemize}

The organization of the rest of the paper is as follows: Section \ref{sec:related} discusses the related works of RGB-D salient object detection and saliency-based image summarization. Section \ref{sec:PROBLEM FORMULATION} presents the formulation of the image summarization problem. In the Section \ref{sec:method}, we present the proposed summarization approach. Experimental results and performance comparison are presented in the Section \ref{sec:results}. The paper concludes in the Section \ref{Conclusion}. 
\tikzstyle{block} = [draw, rectangle, 
    minimum height=3em, minimum width=6em]
\tikzstyle{input} = [coordinate]
\tikzstyle{output} = [coordinate]
\tikzstyle{dotted_block}=[draw, rectangle, line width=1pt, dash pattern=on 1.5pt off 3pt on 6.5pt off 3pt, rounded corners]

\section{Prior Works:}
\label{sec:related}
Image summarization and RGB-D salient object detection are two important aspects of the $3$D scene summarization. 
Hence, this section describes the related works in these categories separately in the following subsections:

\subsection{Image Summarization}

Image summarization has been a focus area of research for the last two-to-three decades \cite{kuzovkin2017context, pasini2022semantic, sharma2022task, mahalakshmi2022summarization, ozkose2019diverse, singh2020image, singh2019image, sreelakshmi2021image, theisen2023motif}. Consequently, numerous methods have been introduced in the literature. In several works, the image summarization problem is formulated as an image collection summarization which aims to select a set of representative images from a large image corpus. Earlier methods utilized low-level visual features and employed clustering \cite{simon2007scene, camargo2009multi, deng2007content, stan2003eid}, similarity pyramids \cite{chen2000hierarchical}, and graph \cite{cai2004hierarchical, gao2005web} techniques. Later, Sinha \etal~\cite{sinha2011summarization} used multidimensional features such as visual, temporal, event type, location, and people. Yang \etal~\cite{yang2013image} formulated the summarization problem using a dictionary learning of a SIFT (Scale Invariant Feature Transform)-Bag of Words model for generating a summary.
Recently, both Singh \etal~\cite{singh2019image} and Sreelakshmi \etal~\cite{sreelakshmi2021image} adopted GoogLeNet \cite{szegedy2015going} as image features extractor and for summarization Singh \etal~\cite{singh2019image} employed Generative Adversarial Networks (GAN) \cite{creswell2018generative}, whereas Sreelakshmi \etal~\cite{sreelakshmi2021image} introduced two unsupervised methods one is using OneClassSVM and another using an autoencoder. Sharma \etal~\cite{sharma2022task} proposed task-specific image corpus summarization using semantic information and self-supervision. The authors used a Resnet-$34$ backbone for feature detection and performed k-means clustering on the semantic spaces for summarization and self-supervision. 
Almost all the methods discussed above are designed for the dataset consisting of $2$D images that do not contain depth information. Hence fails to incorporate the depth information in predicted saliency maps.

Some other works focus on the construction of image summaries by extracting contextual representative information from a single image or video given as input \cite{meena2023review}.
Chen \etal~\cite{chen2012visual} proposed a face-aware spatiotemporal saliency-based summarization method that integrates the detection algorithms of human faces and spatiotemporal attention. Instead of visual saliency, Shi \etal~\cite{shi2009context} utilized context saliency, combining statistical saliency and geometric information, to accurately describe semantic importance under the Bayesian framework. Zhang \etal~\cite{zhang2014saliency} utilized a context-aware saliency guided framework for scene classification. Jiang \etal~\cite{jiang2015salient} utilized a bottom-up approach based on the region's color, depth, layout, and boundary information to produce robust foreground and background salient regions. 
Thomas \etal~\cite{thomas2016perceptual} utilized motion characteristics to calculate visual attention scores.
Deng \etal~\cite{deng2023rgb} introduced a depth stack module and a saliency map refusing module-based method for ranking salient objects in complex indoor scenes. 
Hou \etal~\cite{hou2022multi} proposed a method for multi-modal fusion of $2$D RGB features with density distribution features of the frustum point cloud of an object. However, it fails under different environmental conditions.

Hence, we conclude from the above discussion that $3$D image summarization is rarely addressed. Furthermore, no technology exists that summarizes data via visual saliency guided by volumetric saliency. Therefore, this domain still requires enhanced attention.


\begin{figure*}[!hbt]
 \begin{tikzpicture}

  \node [input, name=input1] {};
 \node [input,  right of=input1,node distance=0.8cm] (input) {};
 
  \node [input,  right of=input,  node distance=1.2cm] (joint) {};
  \node [input,  above of=joint,  node distance=1.4cm] (jointup) {};
  \node [input,  below of=joint,  node distance=1.4cm] (jointbel) {};
 \draw [draw,-] (joint) -- node {} (jointup); \draw [draw,-] (joint) -- node {} (jointbel);
  \node [input,  above of=joint,  node distance=0.5cm] (jointup1) {};
  \node [input,  below of=joint,  node distance=0.5cm] (jointbel1) {};
  
  \node [input,  right of=input,  node distance=2.4cm] (a) {};

  \node [parallelepiped,draw=orange!60, fill=orange!20, above of=a,  node distance=0.5cm, minimum width=1.7cm, align=center] (ac) {Texture};

   \node [parallelepiped,draw=orange!60, fill=orange!20, above of=ac,  node distance=0.9cm, minimum width=1.7cm, align=center] (at) {Color};
   
   \node [input, above of=at,  node distance=0.6cm, label=Visio-Spatio Features] (feature) {};   

   \node [parallelepiped,draw=orange!60, fill=orange!20, below of=a,  node distance=0.5cm, minimum width=1.7cm, align=center] (as) {Spatial};

   \node [parallelepiped,draw=orange!60, fill=orange!20, below of=as,  node distance=0.9cm, minimum width=1.5cm, align=center] (ad) {Directional};

 \draw [draw,->] (jointup) -- node {} (at); \draw [draw,->] (jointbel) -- node {} (ad);
 \draw [draw,->] (jointup1) -- node {} (ac); \draw [draw,->] (jointbel1) -- node {} (as);
   
    \node [parallelepiped,draw=cyan!60, fill=cyan!30, right of=a, node distance=3.1cm, align=center] (b) {\includegraphics[width=.15\textwidth]{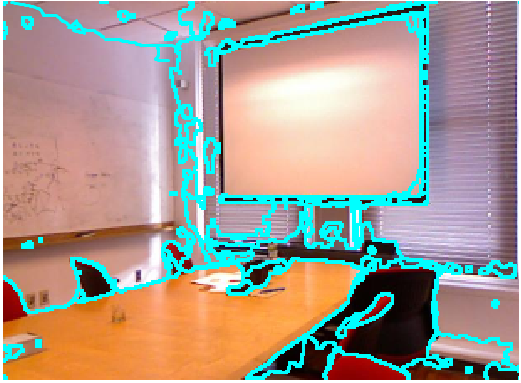}\\Parametric\\ Model-based\\ Segmentation};
 
     \node [parallelepiped, draw=magenta!60, fill=magenta!20, right of=b, node distance=3.4cm, align=center] (c) {\includegraphics[width=.15\textwidth]{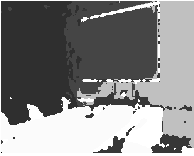}\\ Volumetric\\Saliency \\ Prediction};
     
     \node [parallelepiped, draw=red!50, fill=red!10, right of=c,
        node distance=2.8cm, minimum width=1.8cm, align=center] (d) {Summary\\ Generation};
    \node [output,  right of=d,node distance=1.4cm] (output) {};

\draw [draw,->] (input1) -- node[label={RGB-D Image}]{\includegraphics[width=.15\textwidth]{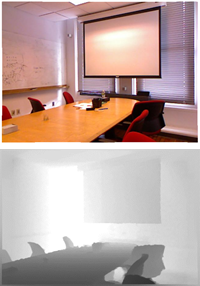}} (input);
 \draw [draw,-] (input) -- node {} (joint);
 
  \node [output,  right of=a,node distance=1.25cm] (djoint) {};
 \node [input,  above of=djoint,  node distance=1.4cm] (djointup) {};
  \node [input,  below of=djoint,  node distance=1.4cm] (djointbel) {};
 \draw [draw,-] (djoint) -- node {} (djointup); \draw [draw,-] (djoint) -- node {} (djointbel);
  \node [input,  above of=djoint,  node distance=0.5cm] (djointup1) {};
  \node [input,  below of=djoint,  node distance=0.5cm] (djointbel1) {};
\draw [draw,-] (at) -- node {} (djointup); \draw [draw,-] (ad) -- node {} (djointbel);
\draw [draw,-] (ac) -- node {} (djointup1); \draw [draw,-] (as) -- node {} (djointbel1);

 \draw [draw,->] (djoint) -- node {} (b);
\draw [draw,->] (c) -- node {} (d);
 \draw [draw,->] (b) -- node {} (c);
 \draw [draw,->] (d) -- node[pos=0.8,right, label={Summary}] {\includegraphics[width=.15\textwidth,height=.16\textwidth]{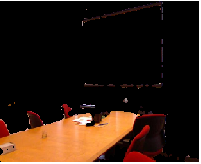}} (output);

\node [below right= 2.1cm and 0.9cm of djoint, align=center] (g) {\textbf{(a) Work-flow of Proposed Method}};


\node [dotted_block, draw=black!40, fit = (input) (a) (b) (c) (d) (output), inner xsep=24mm, inner ysep=9mm, xshift=5.5mm] (aa) {};

\node [input, below left=4.4cm and 0.6cm of input, align=center] (ip) {};
\node [input, below left=4.4cm and 0.6cm of input, align=center] (sa1) {};
\draw [draw,->] (ip) -- node[label=below:{Segmented Image}] {\includegraphics[width=.1\textwidth, height=.16\textwidth]{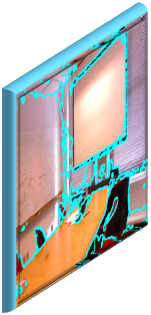}} (sa1);

\node [input, right of=sa1, node distance=0.96cm, align=center] (sa) {};


\node [input, right of=sa, node distance=1.9cm, align=center] (sb) {};
\draw [draw,-] (sa) -- node [above] {Point cloud of} node [below] {each segment} (sb);

\node [output, right of=sb, node distance=1.4cm] (op) {};
\draw [draw,->] (sb) -- node {} (op);

\node [output, right of=op, node distance=2.5cm] (opv) {};

\node[parallelepiped, draw=black!60, fill=lightgray, below of=opv, node distance=2.3cm, align=center] (opb) {\includegraphics[width=.25\textwidth,height=.32\textwidth]{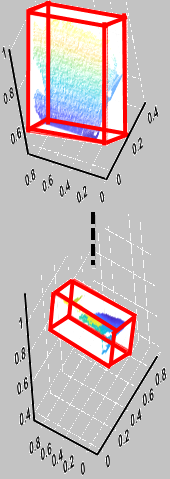}\\$3$D OBB Detection\\for Each Segment};

\node [output, below of=opb, node distance=1.3cm] (opbv1) {};
\node [output, right of=opbv1, node distance=0.8cm] (opb1) {};

\node [output,left of=opb1, node distance=3.1cm] (opb11) {};

\node [parallelepiped, draw=teal!60, fill=teal!20, below of=sb, node distance=3.62cm, align=center] (sc) {\\Volume-based\\Saliency Score\\Computation\\}; \draw [draw,->] (opb11) -- node {} (sc);

\node [output, left of=sc, node distance=1.8cm] (sam1) {};
\draw [draw,->] (sc) -- node {} (sam1);
\node[left of=sam1, node distance=1.0cm, label=below:{Saliency Map}] (sam) {\includegraphics[width=.1\textwidth,height=.16\textwidth]{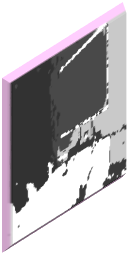}} (sam) ;

\node [below right= 2.6cm and 0.3cm of sam1, align=center] (sam2) {\textbf{(b) Volumetric Saliency Prediction}};

\node [dotted_block, draw=black!40, fit = (ip) (sc) (sam2), inner xsep=19.5mm, inner ysep=9mm, xshift=6.5mm, yshift=6mm] (bb) {}; 
\node [input, below right=4.4cm and 0.28cm of output, align=center] (ip2) {};
\node [left of=ip2, node distance=1.0cm, align=center] (sa2i) {};

\node [right of=sa2i, node distance=2.2cm, align=center, rotate=3] (sa2ii) {\includegraphics[width=.1\textwidth, height=.16\textwidth]{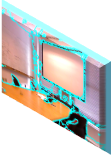}};

\node [input,below of=sa2ii, node distance=1.3cm, align=center] (sa21p) {};
\node [input,left of=sa21p, node distance=2.4cm, align=center] (sa21) {};
\draw [draw,->] (sa2ii) -- node [above, rotate=24]{\footnotesize{if $\mathcal{S}_V\{O_i^k\} > \tau$}} (sa21);
    
\node [input, below of=sa2i, node distance=2.1cm, align=center] (sa2) {};
\node [input, below of=sa21, node distance=0.8cm, align=center] (sap2) {};

\node[below of=sa2, node distance=1.5cm] (sb2) {\includegraphics[width=.38\textwidth,height=.28\textwidth]{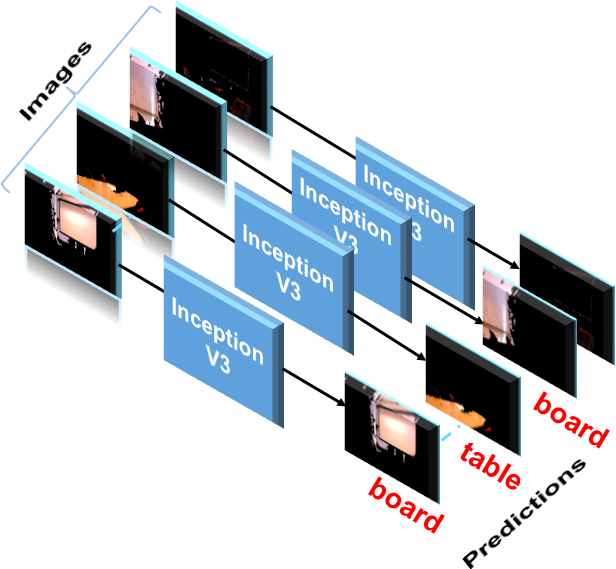}} ;

\node [output, below of=sb2, node distance=3.0cm, align=center] (se2) {};

\node [below of=se2, node distance=0.28cm] (sf2) {\textbf{(c) Object Classifaction}};

\node [dotted_block, draw=black!40, fit = (ip2) (se2) (sf2), inner xsep=17mm, inner ysep=7mm, xshift=1.2mm, yshift=7.8mm] (cc) {};

\end{tikzpicture}
\vspace{-0.5cm}
\caption{Proposed volumetric saliency guided summarization method.}
 \label{fig:method}
\end{figure*}
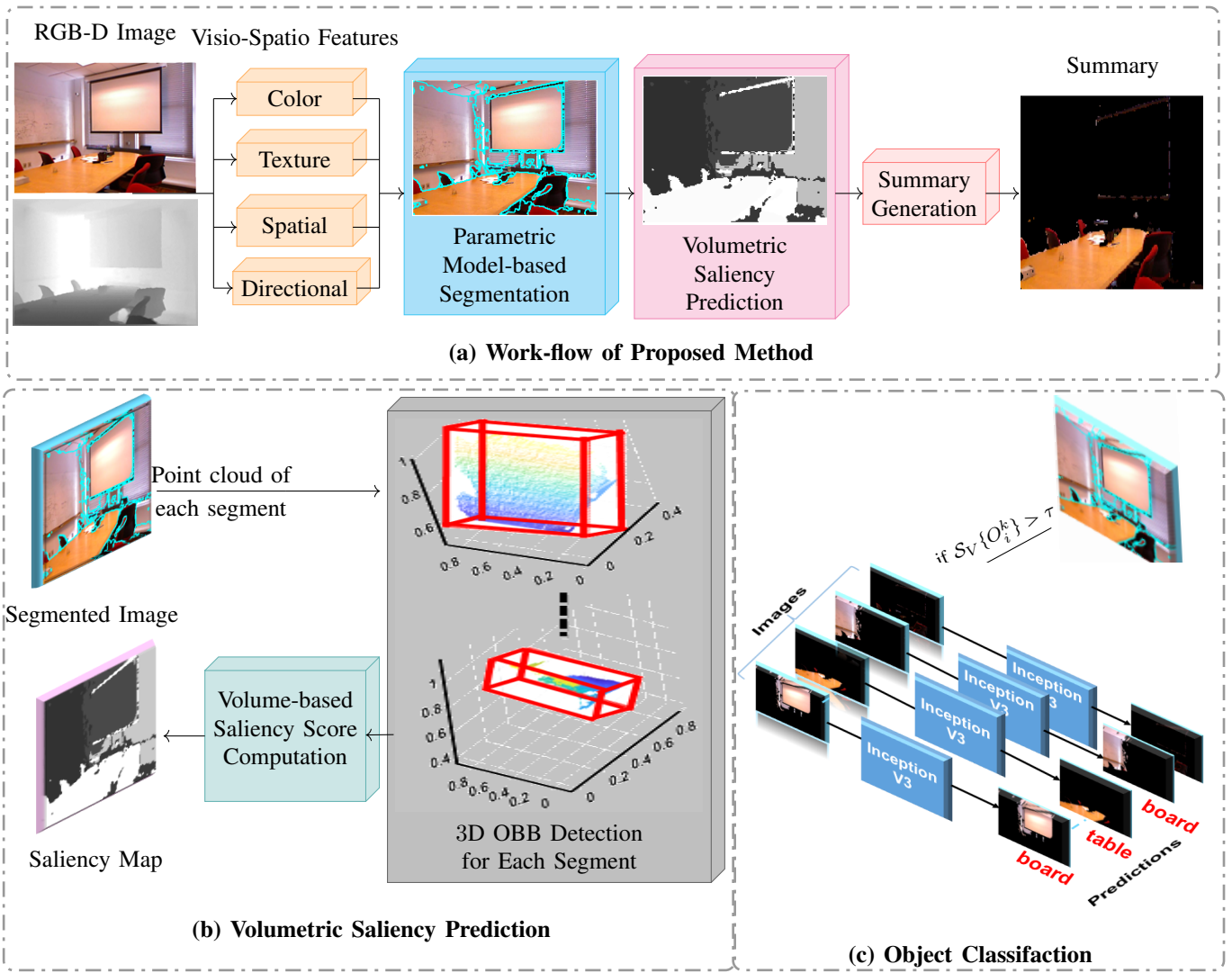

\subsection{RGB-D Salient Object Detection}
Conventional RGB-D salient object detection methods extracted the hand-crafted features from RGB and depth images and fused them as in \cite{lang2012depth, fan2014salient, peng2014rgbd} for saliency generation. However, this results in limited generalizability for saliency generation in the case of complex scenes. Recently, with the development of deep learning techniques, many deep learning-based fusion methods for RGB-D salient object detection achieved significant success e.g. \cite{fan2020rethinking, cong2019going, lou2020exploiting, pang2020hierarchical, huang2020joint, li2020rgb, zhao2021self, zhang2021uncertainty, yang2022bi, ji2022dmra}. 
Piao \etal~\cite{piao2019depth} suggested a depth-induced multi-scale recurrent attention network and designed a depth residual block to integrate cross-modal features. Chen \etal~\cite{chen2019three} adopted a three-stream attention-aware architecture to explore cross-modal and cross-level complements. Liu \etal~\cite{liu2020learning} utilized a self-mutual attention mechanism inspired by the non-local model \cite{wang2018non} for salient object detection. Fu \etal~\cite{fu2021siamese} utilized the Siamese network to jointly learn RGB and depth images to extract useful complementary features. Chen \etal~\cite{chen2021rgb} proposed the first $3$D CNNs-based method for RGB-D salient object detection. Fan \etal~\cite{fan2020rethinking} proposed a deep depth-depurator network composed of three branches, which can filter redundant information and learn multi-modality feature information for salient object detection. Ji \etal~\cite{ji2021calibrated} suggested a depth calibration and fusion network to deal with the noise in the depth images. Similarly, Li \etal~\cite{li2021hierarchical} adopted a Hierarchical Alternate Interaction Network (HAINet) to minimize noise in the depth maps for efficient cross-modal interactions. Zhang \etal~\cite{zhang2021uncertainty} introduced the uncertainty-inspired probabilistic RGB-D saliency prediction via conditional variational autoencoder. Zhu \etal~\cite{zhu2023s} introduced a self-supervised self-ensembling network for semi-supervised RGB-D salient object detection by leveraging the unlabeled data and exploring a self-supervised learning mechanism. 
In these methods, all the salient objects are not completely detected, or parts of the backgrounds are not well suppressed in the predicted saliency maps. This may be due to the inaccuracy in the delineation of the boundary of salient objects. To improve object boundary detection, we introduce a two-stage probabilistic fusion framework of geometric and visual features.

Recently, Zhang \etal~\cite{zhang2022c} introduced a Model-specific Dynamic Enhanced Module that dynamically enhances the intra-modality features of RGB and depth with global context guidance. Whereas Li \etal~\cite{li2023mutual} introduced an inter-modal mutual information regularization-based method to explicitly model the contribution of RGB and depth for weakly-supervised RGB-D saliency detection. For multi-modality-based salient object detection, Chen \etal~\cite{chen2022modality} introduced a modality-induced transfer-fusion network (MITF-Net). Wu \etal~\cite{wu2023hidanet} introduced a Hierarchical Depth Awareness network (HiDAnet) for salient object detection that leverages fine-grained details and merges them with semantic cues through the local channel attention. Sun \etal~\cite{sun2023catnet} proposed a cascaded and aggregated Transformer Network (CATNet) to aggregate multi-scale features for salient object detection. The method successfully suppresses noise in low-level features and mitigates the differences between features at different scales. Zhang \etal~\cite{zhang2023feature} utilized a calibration-then-fusion-based two-stage model to suppress the influence of two types of depth images (\ie low-quality ones and foreground-inconsistent ones) on final saliency prediction. Using this model, foreground objects are predicted to be salient with more complete boundaries and less disturbing backgrounds. However, the background objects are predicted to be non-salient. Thus, the model fails for indoor scenes saliency detection task that also contains the salient objects in the background. 


Almost all the aforementioned works considered the foreground as a salient area. These works can learn high-level representations to explore complex correlations between RGB images and depth cues for improving salient object detection performance. However, these methods are unsuitable for indoor scene classification tasks due to the absence of scene-characterizing objects in the background.

In literature, a few works focused on salient areas in foreground and background jointly \cite{jiang2015salient, ren2015exploiting, liang2018stereoscopic}. Recently, Zhang \etal~\cite{zhang2021bilateral} proposed the bilateral attention mechanism, including foreground-first and background-first attention. Despite considering both foreground and background, it still fails to detect all scene-characterizing objects. Moreover, it should be noted that the high performance of the method is at the cost of a complex architecture, which may limit its application. Zeng \etal~\cite{zeng2023airsod} developed a lightweight architecture to solve this issue. Although it provides good performance, but it performs poorly in scenarios in which the depth difference between the background and salient object is not significant.

Thus, we observe from the current literature that there does not exist any optimal saliency detection methodology that includes both background and foreground scene parts effectively. Also, visual and depth features are different modalities; hence, direct fusion of these modalities cannot be an effective strategy. However, one modality can be utilized to effectively guide the selection of other modalities. In this paper, we utilize volumetric information to guide salient object detection using visual features.

\section{Problem Formulation}
\label{sec:PROBLEM FORMULATION}
The scene classification can be performed effectively using a scene-defining summary. Thus, we define our problem statement as: {\it{a saliency estimation approach for effective image summarization for the RGB-D indoor scene classification.}} The problem statement is formulated as follows: \\

Let a set of RGB-D images $I$ in the dataset $\mathcal{D}$ belonging to the scenes of $\mathcal{N}$ different categories $\{C_i\}_{i=1:\mathcal{N}}$. Each scene category $C_i$ is a collection of $T$ characterizing objects $C_i = \{O_{1}^i, O_{2}^i, ..., O_{T}^i\}$ such that $(C_i = C_j) \iff (i=j)$. If summary $\{O_i\}_{i=1:\mathcal{M}}$ ($\mathcal{M} \gtreqless T$) of the scene from an image $I_k \in C_k$ is estimated through a visual saliency detection function $\mathcal{S}_v(.)$ guided by the volumetric saliency $\mathcal{S}_V(.)$, then a classifier $\mathcal{C}$ is required  such that 

\begin{eqnarray}
    \text{Find} \  \mathcal{S}_v(.)  &\ &  \text{s.t.} \ \ \mathcal{S}_v(I_k) = \{O_i^k\}_{i=1:\mathcal{M}}  \\  \nonumber
    \text{and}&& \ \  \mathcal{S}_V \{O_i^k\}_{i=1:\mathcal{M}} > \tau\\  \nonumber
    \text{and}&& \ \  \sum_{\mathcal{K} \in \mathcal{O} } ||\mathcal{C}(I\{O_i^k\}) - \mathcal{K}||^2  \quad \text{is minimal.}
    \label{eqn:problem}
\end{eqnarray}

Where $\tau$ denotes a threshold and each object from $\{O_i^k\}_{i=1:\mathcal{M}}$ is considered as a new image $I\{O_i^k\}$. 
$\mathcal{C}(I\{O_i^k\})$ is the object category predicted by a classifier $\mathcal{C}$ and $\mathcal{K}$ is the object's true category.
In this work, a summary generation task includes an Inceptionv$3$ \cite{szegedy2016rethinking} model as classifier $\mathcal{C}$ that was trained on total $\mathcal{O}=36$ object categories.

For the scene classification task, 
an SVM classifier represented by $\mathcal{C}$ is selected 
for performance evaluation of the generated summaries using saliency detection function $\mathcal{S}_v(.)$. To accomplish high efficacy, it is required that
\begin{eqnarray}
    \sum_{k\in \mathcal{D} } ||\mathcal{C}(\mathcal{S}_v(I_k)) - k||^2  \quad \text{is minimal.}
    \label{eqn:problem1}
\end{eqnarray}



\section{Proposed Volumetric Saliency Guided Image Summarization}
\label{sec:method}

The proposed summarization framework shown in Fig. \ref{fig:method} comprises three major stages, namely \textit{Visio-Spatio Feature Extraction}, \textit{Parametric Model-based Segmentation}, and \textit{Volumetric Saliency Guided Summary Generation}. Further description of each of these stages is explained in the subsequent subsections.

\subsection{Visio-Spatio Feature Extraction}

This subsection discusses the methodology to extract essential features for all the objects of an image for summarization purposes. Therefore, extracting all necessary features is a crucial step in effective saliency generation for the summary of indoor RGB-D scenes. In this regard, both spatial and visual features are vital. We utilize color and texture-related features as visual features for saliency detection. LAB color space is utilized as color features. In addition, texture features are also extracted using the Gray level co-occurrence matrix (GLCM) \cite{haralick1973textural}. The extracted GLCM features are correlation, energy, information measure of correlation, homogeneity, dissimilarity, and entropy. These texture features provide high discrimination accuracy while requiring less computation time; consequently, they are useful for segmentation tasks.

As visual features are unreliable under some indoor environmental conditions (\eg various effects caused by spatially varying illumination and the presence of shadows), incorporating spatial features is necessary to provide a better representation. Spatial ($3$D) location and directional features are used as spatial features for saliency detection. The $3$D position of the pixel is taken as spatial features using point-cloud representation, i.e., $PC(X, Y, Z)$. The position $X = (r - {O_{x}})  (\frac{Z}{f_{x}})$, $Y = (c - {O_{y}})  (\frac{Z}{f_{y}})$, and $Z = g(r,c)$ at pixel location $(r,c)$ with depth map $g(r,c)$ is computed using \textit{Thin Lens} model. Here $({O_{x}}$ and ${O_{y}})$ are the image center, ${f_{x}}$ and ${f_{y}}$ denote the horizontal and vertical focal lengths of the camera. The ($3$D) points often fail to provide sufficient information to delineate the intersections of planar surfaces with different orientations, such as walls, floors, ceilings, etc. Therefore, directional features are included. Directional features (\ie surface normal) are computed using the method given in \cite{hoppe1992surface}. 
The computed color features vector is of dimensionality $3$ that contains the LAB components. Similarly, the spatial and directional features are also of dimensionality $3$. 

For each pixel, the color, spatial, and directional features are concatenated to form a feature vector of dimensionality $9$, which will be utilized for further processing, as discussed in the subsequent subsection. 
\subsection{Parametric Model-based Segmentation}
A visual summary of a scene from an image $I$ is a collection of salient objects $\{O_i\}_{i=1:\mathcal{M}}$. Thus, the segmentation is performed using extracted feature vectors in the previous subsection for detecting the objects in the scene. For this purpose, we utilize superpixel segmentation followed by region merging. Superpixel segmentation initially segments the image into a set of regions/segments. Then, the region-merging refines this set by merging regions that are susceptible to over-segmentation. A detailed description of each stage is given in subsequent subsections.

\subsubsection{SLIC with Additional Spatial Features}
In this work, we utilize the SLIC (Simple Linear Iterative Clustering) algorithm, which is a standard segmentation algorithm for segmenting out the objects $O_i^k$ in the given RGB-D image $I_k$ of the scene. The SLIC  algorithm \cite{achanta2012slic} generates superpixels by clustering pixels based on the distance (similarity) between the color vector $(L, a, b)$ and the image pixel coordinates $(r, c)$. However, we incorporate additional $3$D location (\ie spatial coordinates) and directional features terms to include $3$D spatial features effectively. As a result, the new distance measure $D(.)$ for pixels $(i$ and $j)$ in image $I$ is computed as given in \ref{eq:direcslic}.

\begin{equation}
D(i,j) = \frac{d_{ij}^{c}}{m} + \frac{d_{ij}^{p}}{S} +  \frac{{d_{ij}^{s}}}{a} + \frac{d_{ij}^{\theta}}{b} + \frac{d_{ij}^{\alpha}}{d} 
\label{eq:direcslic}
\end{equation}

Where $S$ denotes the grid interval of superpixels; $m$ denotes the compactness factor; ${d_{ij}^{c}}$, ${d_{ij}^{p}}$, and ${d_{ij}^{s}}$ are Euclidean distances between color, image pixel position, and spatial location, respectively; the ${d_{ij}^{\theta}}:= \left|{\theta_{j}}-{\theta_{i}} \right|$ and ${d_{ij}^{\alpha}}:= \left|{\alpha_{j}}-{\alpha_{i}} \right|$ are the distances between directional components (\ie azimuth and elevation angles) for point cloud of $(i$ and $j)$ respectively; $a$, $b$ and $d$ respectively denote the maximum distances of the spatial location, local azimuth angle, and elevation angle respectively. 

\begin{figure}[!hbt]
    \centering
    \includegraphics[width=0.96\linewidth]{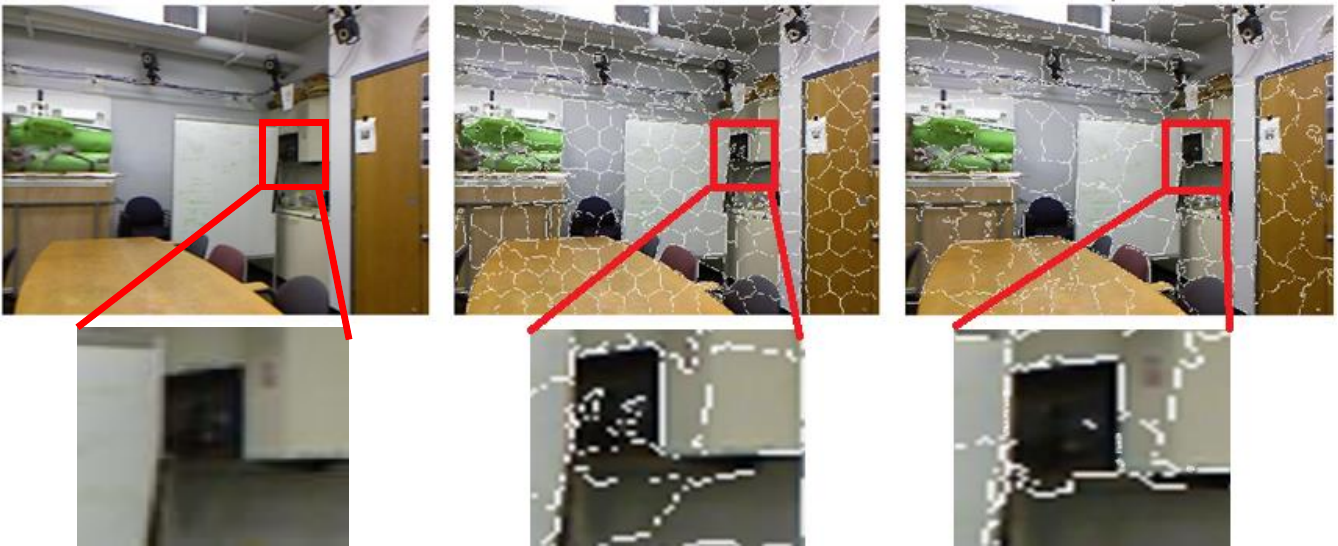}
    \centerline{\footnotesize{(a) \hspace{0.85in} (b) \hspace{0.85in} (c)}}
    \caption{Segmentation comparison of modified SLIC using additional spatial features with original SLIC. (a) RGB  image, (b)-(c) segmentation results using original SLIC \cite{achanta2012slic}, and modified SLIC \ref{eq:direcslic} with additional spatial features, respectively.}
    \label{fig:sliccomp}
\end{figure}

Figure \ref{fig:sliccomp} shows an example to compare the segmentation performance of SLIC with additional spatial features with the original SLIC algorithm \cite{achanta2012slic}. The figure shows that the original SLIC \cite{achanta2012slic} fails to distinguish between object and shadow. In contrast, improved SLIC with additional spatial features demonstrates better efficacy with numerous effects caused by spatially varying illumination. To illustrate the quantitative performance, BDE \cite{harouni2018color}, and VOI \cite{harouni2018color} metrics are compared in Tab. \ref{tab:sliccompquan} on the NYUv2 \cite{silberman2012indoor} dataset. The table shows that the SLIC with additional spatial features achieves $0.5$ and $1.7$ score improvement in terms of BDE and VOI, respectively, on scores of the original SLIC. Though the modified superpixel segmentation method (\ie SLIC with additional spatial features) effectively detects the boundaries of the segments, this superpixel segmentation may result in over-segmentation. 
Thus, it is required to merge these over-segmented regions to achieve good segmentation results, as discussed in the following subsection.

\begin{table}[!hbt]
\centering
\caption{Quantitative comparison between SLIC \cite{achanta2012slic} and Modified SLIC \ref{eq:direcslic}.}
\label{tab:sliccompquan}
\renewcommand{\arraystretch}{1.2}
\begin{tabular}{c|c|c}
\hline
Metrics & SLIC \cite{achanta2012slic} & \begin{tabular}[c]{@{}l@{}} Modified SLIC\end{tabular} \\  \hline
BDE \cite{harouni2018color} {($\downarrow$)} & $6.2$ & $5.7$\\ 
\hline
VOI \cite{harouni2018color}  {($\downarrow$)} & $5.9$ & $4.2$ \\ \hline
\end{tabular}
\end{table}

\subsubsection{Over-segmentation Error Reduction}

Over-segmentation drastically affects the performance of the scene summarization using objects $O_i^k$ in the scene. One main reason for over-segmentation is the random variation of feature values. Thus, a probabilistic framework is adopted for region merging by taking a mixture model for probability distribution functions \cite{kasarapu2015minimum}, \cite{banerjee2005clustering}. We consider the 'multivariate Gaussian distribution' \cite{kasarapu2015minimum} for $3$D location and the 'multivariate Fisher distribution' \cite{banerjee2005clustering} for surface normal.

\begin{figure}[!hbt]
    \centering
    \includegraphics[width=0.98\linewidth]{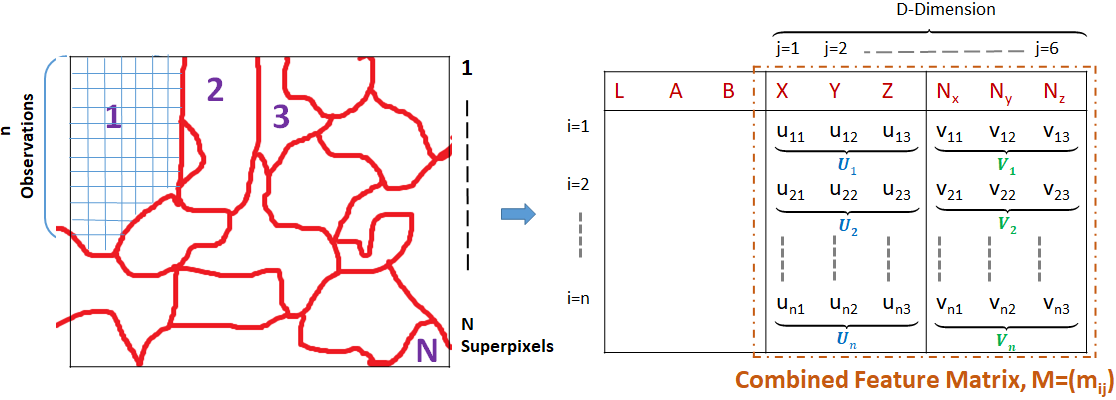}
    \centerline{\footnotesize{Superpixel Boundary Mask \hspace{0.28in} Feature Matrix for Superpixel $1$.}}
    \caption{A superpixel example and feature vector.}
    \label{fig:emdiag}
\end{figure}

Let for a set of $N$ superpixels obtained through the previous stage, $\overrightarrow{U_{i}}$ and $\overrightarrow{V_{i}}$ for $i=1,2,...,n$ be a $3$-dimensional observed $3$D location and direction (surface normal) of $n$ observations in each superpixel as shown in Fig. \ref{fig:emdiag}. Then, the combined feature matrix that follows the \textit{Fisher–Gaussian }distribution \cite{wang2021matrix} is given as $
\textbf{M}= [{\overrightarrow{U_{1}},\overrightarrow{U_{2}},....,\overrightarrow{U_{n}}}, {\overrightarrow{V_{1}},\overrightarrow{V_{2}},....,\overrightarrow{V_{n}}}]
$.
The joint density function ${f_{c}(m_{i}|\Theta_c)}$ for a mixture of the \textit{Fisher-Gaussian} random variable in each superpixel can be written as ${f_{c}(m_{i}|\Theta_c)} = \sum_{c=1}^{K} \pi_c \times f_{c}(U_{i}|\Theta_c) \times f_{c}(V_{i}|\Theta_c)$. 
Where $c=1,2,.., K$ is the number of components in the mixture model. The pdfs ${f_{c}(U_{i}|\Theta_c)}$ and ${f_{c}(V_{i}|\Theta_c)}$ are the density functions for multivariate GMM and Fisher distribution, respectively. And $\pi_c$ is called the mixing coefficient. 
To estimate optimal parameters $\Theta_{c}=\{\mu_{c}, \Sigma_{c}, \eta_{c}, \kappa_{c}, \pi_c\}$ of the mixture model, we have utilized the \textit{Expectation Maximization} (EM) algorithm \cite{moon1996expectation}. Here $\mu_{c}$ and $\Sigma_{c}$ represent the mean vector and covariance matrix of multivariate GMM. The parameters $\eta_{c}$ and $\kappa_{c}$ represent mean and concentration parameters, respectively, in multivariate Fisher distribution.

These estimated optimal parameters are then used for segment merging. In addition, we also include a parameter called curvature that defines the planarity of the regions. The main goal of merging is to merge the adjacent similar regions that are susceptible to over-segmentation. The similarity between two superpixels is defined by the similarity between different parameters associated with each region as given in \ref{eqn:similarity}. 


\begin{equation}
M_{ij}= 
 \begin{cases}
    1; & \hspace{0.1ex} \textrm{if }( \kappa, c_{u}) > ({k_{th}}, {c_{th}}),\\  
       & \hspace{0.1ex}  ({D_{T}(i,j)},  {d_{w}(i,j)}) < ({D_{th}}, \delta)  \\
    0;                 & \hspace{0.1ex} \textrm{Otherwise}
  \end{cases}
\label{eqn:similarity}
\end{equation}

The equation indicates that if two adjacent segments $i$ and $j$ belong to a planar region (\ie $\kappa>k_{th} \text{ and } c_{u}>c_{th}$) and if the distances ${D_{T}(i,j)}$ and ${d_{w}(i,j)}$ between them is less than thresholds ${D_{th}}$ and $\delta$ then,  they are said to be similar. Where, ${D_{T}(i,j)}$ is computed as the distance between texture components; ${d_{w}(i,j)}$ is expressed as: 
\begin{equation}
{{d_{w}(i,j)}=\frac{\beta_{1} \ast {D_{c}(i,j)} + \beta_{2} \ast {D_{de}(i,j)} + \beta_{3} \ast {D_{SN}(i,j)}} {\beta_{1}+\beta_{2}+\beta_{3}}} 
\end{equation}
Here, ${D_{c}(i,j)}$ denotes the Euclidean distances between the color of segments $i$ and $j$. Whereas, ${D_{de}(i,j)}$ and ${D_{SN}(i,j)}$ are the distances between the density functions and computed by using the KL divergence \cite{duchi2007derivations} between the density functions of segments (\ie superpixel) $i$ and $j$. $c_{u}$ is the curvature that measures the degree of bending of a surface, and $\beta_{1}, \beta_{2}, \text{ and } \beta_{3}$ are the weights.
To determine these weights and optimal threshold values (i.e., $c_{th}$, $D_{th}$, and $\delta$), we evaluated the performance of the parametric model-based segmentation approach (the second stage of proposed summarization framework) in terms of quantitative measures BDE\cite{harouni2018color}. We computed BDE\cite{harouni2018color} metrics for different combinations of weights and threshold values. Next, the combination of weights and thresholds that offer the lowest BDE \cite{harouni2018color} values are selected as optimal. The thresholds that we obtained by above experiment are ${c_{th}=0.05}, {k_{th}=5}$, ${D_{th}=1.5}$, and ${\delta= 0.35}$. and the weights are ${\beta_{2}=0.6}$, ${\beta_{1}=1-\beta_{2}}$, and ${\beta_{3}=0.5}$.

\begin{figure}[!t]
    \centering
    \includegraphics[width=0.96\linewidth]{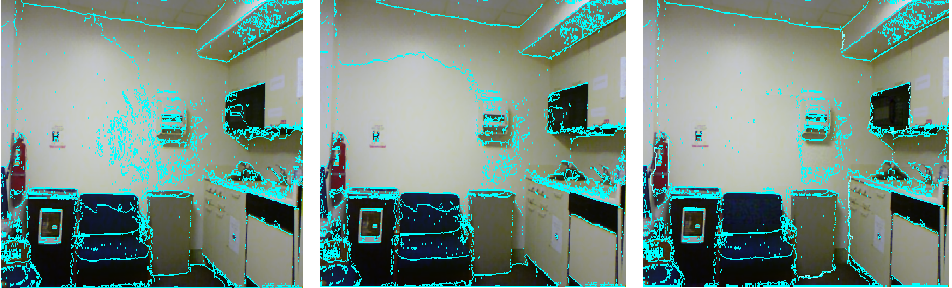}
    \centerline{\footnotesize{(a) \hspace{0.85in} (b) \hspace{0.85in} (c)}}
    \caption{Region merging based on (a) depth only, (b) depth \& direction only, and (c) multiple parameters as in \ref{eqn:similarity}.}
    \label{fig:slic}
\end{figure}

Furthermore, to illustrate the effectiveness of region merging using  \ref{eqn:similarity}, we performed an ablation study by removing and replacing the parameters in  \ref{eqn:similarity}. We started with the region merging based on depth only (\ie ${D_{de}(i,j)}$). This can be done by removing all the parameters from \ref{eqn:similarity} except $d_w(i,j)$. At the same time modifying $d_w(i,j)$ by removing ${D_{c}(i,j)}$ and $D_{SN}(i,j)$. Fig. \ref{fig:slic}(a) shows an example of generated output based on depth only. We observed from the figure that the segmented image includes the regions that present at the same depth but belong to different segments. Moreover, it over-segments the planar regions that have varied depth values. Next, the merging based on depth and direction is adopted to see the benefit of including surface normal. In this case, only ${D_{c}(i,j)}$ is removed from $d_w(i,j)$, and the other setting remains similar to the first case (merging based depth only). Figure \ref{fig:slic}(b) shows an example indicating that the performance of region merging based on depth and direction is improved compared to depth only. However, the regions belonging to different segments but are in the same direction (\ie having the same orientation) are also merged. Therefore, we applied multiple parameters to compute the region's similarity to address the above limitations. As a result, we performed the region merging based on \ref{eqn:similarity} as shown in Fig. \ref{fig:slic} (c). As illustrated in Fig. \ref{fig:slic}, region merging using \ref{eqn:similarity} generates an optimal result compared to other cases. In addition, we observe that it successfully delineates the different regions or segments within the image.

Further, these extracted segments belonging to objects $O_i^k$ are used as input to the next stage of the proposed summarization framework \ie volumetric saliency guided summary generation.

\subsection{Volumetric Saliency Guided Summary Generation}
\label{Saliency map and summary generation}

In this subsection, we aim to create a saliency map $\mathcal{S}_v(I_k) = \{O_i^k\}_{i=1:\mathcal{M}}$ that is useful in the generation of an image summary. We have utilized the segmented objects $O_i^k$ obtained through the parametric model-based segmentation approach. And a saliency map is generated by computing the saliency score for the obtained segments of given images $I$.
Since visual saliency alone is not sufficient for extracting the scene-defining summary. Thus, the volumetric saliency guide is utilized to determine the visual saliency order.

The $3$D points (point cloud) data for each segmented object are given as inputs to find a $3$D oriented bounding box (OBB) \cite{chang2011fast} of the segmented objects $O_i^k$ with the minimum volume for volume computation. Firstly, the convex hull of these points is estimated. Next, the SVD (singular value decomposition) \cite{gene1996matrix} of the $3$D points on the convex hull is employed to extract the axis of the bounding box. Finally, each segment's volume enclosed by OBB \cite{chang2011fast} is used as a saliency score $\mathcal{S}_V\{O_i^k\}$ to generate a volumetric saliency map. The visual saliency guided by the volumetric saliency is obtained using  \ref{eq:mask}.

\begin{equation}
\label{eq:mask}
\mathcal{S}_v(I_k)= 
  \begin{cases}
  0; & \textrm{if } \mathcal{S}_V\{O_i^k\} < \tau, \mathcal{C}({I\{O_i^k\}) \in (\rho)}\\  
1;                 &  \textrm{Otherwise}
  \end{cases}
\end{equation}

Where $\rho$ represents the segments belonging to the wall and floor, respectively.
The equation shows that the objects $O_i^k$ with large volumes are taken to extract a stationary scene-defining summary. Thus, the saliency map retains only those segmented objects $O_i^k$ for which the volumetric saliency score $\mathcal{S}_V\{O_i^k\}$ is greater than a threshold $\tau$. The threshold is set to an experimentally determined value of $\tau=0.2$. For $\tau$ value computation, an experiment is performed by setting the human-annotated summary as groundtruth. Then, the generated summary is compared with the ground truth using the metric F-measure \cite{Saliency-Evaluation-Toolbox}. The threshold value that results in the highest F-measure is taken as a $\tau$ in this paper. 

However, the segments extracted by considering only the constraint $\mathcal{S}_V\{O_i^k\} > \tau$ include some unwanted segments that are not desirable to produce a good summary \eg wall and floor. Therefore, a classifier $\mathcal{C}$ called Inceptionv$3$ \cite{szegedy2016rethinking} model which provides  minimum $\sum_{\mathcal{K} \in \mathcal{O} } ||\mathcal{C}(I\{O_i^k\}) - \mathcal{K}||^2$
is utilized to remove these unwanted segments. This model is trained on the database containing $5400$ images with $36$ object categories $\mathcal{O}$. These images are collected from various online sources. Additionally, data augmentation is used to avoid overfitting.

For unwanted segments (\ie $\rho$) removal and to extract the desirable objects $\{O_i^k\}_{i=1:\mathcal{M}}$, each segment with $\mathcal{S}_V\{O_i^k\} > \tau$ is considered as a new image $I\{O_i^k\}$ by masking and zero padding the original RGB image. Then, each of these images is given to Inceptionv$3$ \cite{szegedy2016rethinking} that classifies the given image as shown in Fig. \ref{fig:method}(c). The trained Inceptionv$3$ \cite{szegedy2016rethinking} model produces classification results with $90.5\%$ accuracy. The results of the classifier $\mathcal{C}(I\{O_i^k\})$ are used to modify the volumetric saliency map. If image $I\{O_i^k\}$ for which the class $\mathcal{C}(I\{O_i^k\}) \in \rho$, the corresponding segment is discarded. 
Finally, the volumetric saliency map is converted into a binary mask $\mathcal{S}_v(I_k)$ by assigning $0$ to the pixels belonging to these unwanted segments and segments with $\mathcal{S}_V\{O_i^k\} < \tau$. Whereas pixels belonging to the other segments are set to $1$. The visually attended areas indicated by the obtained mask $\mathcal{S}_v(I_k)$ can be considered as ROI having salient objects $\{O_i^k\}_{i=1:\mathcal{M}}$.


\begin{table*}[!hbt]
\centering
\caption{F-Measure, E-Measure, S-Measure, and MAE comparison of proposed work with SOTA methods.}
\label{tab:mos}
\renewcommand{\arraystretch}{1.3}
\begin{tabular}{l|l|ccccccc}
\hline
\multirow{2}{*}{\begin{tabular}[c]{@{}l@{}}Dataset \\(\# \text{Images})\end{tabular}} & Metrics & \multicolumn{7}{c}{Methods}\\ \cline{3-9}
{}& {} & \cite{lou2020exploiting} & \cite{cong2019going} & \cite{fan2020rethinking} & \cite{zhang2022c} & \cite{zhang2021bilateral} & \cite{li2023mutual} & Proposed Method\\ 
\hline
\multirow{2}{*}{\begin{tabular}[c]{@{}l@{}}NYUv2 \footnote{\url{https://cs.nyu.edu/~silberman/datasets/nyu_depth_v2.html}}\\ ($100$)\end{tabular}} & F-Measure ($\uparrow$) & $0.477$ & $0.690$  & $0.711$ & $0.497$ & $0.683$ & $0.662$ & $\textbf{0.90}$ \\
 & E-Measure ($\uparrow$)  & $0.395$ & $0.733$ & $0.731$ & $0.554$ & $0.728$ & $0.678$ & $\textbf{0.892}$ \\ 
 & S-Measure ($\uparrow$) & $0.470$ & $0.661$ & $0.661$ & $0.530$ & $0.670$ & $0.635$ & $\textbf{0.878}$ \\ 
 & MAE {($\downarrow$)} & $0.371$ & $0.180$ & $0.179$ & $0.230$ & $0.181$ & $0.186$ & $\textbf{0.074}$\\ 
 \hline
\multirow{2}{*}{\begin{tabular}[c]{@{}l@{}}SUNRGB-D\\ \cite{song2015sun} ($145$)\end{tabular}} & F-Measure ($\uparrow$) & $0.461$ & $0.687$  & $0.518$ & $0.536$ & $0.570$ &  $0.567$ & $\textbf{0.757}$ \\
 & E-Measure ($\uparrow$) & $0.431$ & $0.674$ & $0.510$ & $0.551$ & $0.545$ & $0.521$ & $\textbf{0.728}$ \\ 
 & S-Measure ($\uparrow$) & $0.462$ & $0.661$ & $0.463$ & $0.522$ & $0.534$ & $0.543$ & $\textbf{0.708}$\\ 
 & MAE {($\downarrow$)}   & $0.364$ & $0.190$ & $0.307$ & $0.281$ & $0.253$ & $0.246$ & $\textbf{0.166}$ \\ 
 \hline
\end{tabular}
\end{table*}

\section{Results and Discussion}
\label{sec:results}
This section presents the performance evaluation of the proposed volumetric saliency guided image summarization method. Our test setup contains {\it{MATLAB}} {\it{R2022a}} in {\it{Windows 10}} environment with an Intel Xeon CPU and $64$ GB RAM. The NYUv2 \cite{silberman2012indoor}, SUNRGB-D \cite{song2015sun} and Hypersim \cite{roberts2021hypersim} datasets are used for evaluation. The number of images considered for evaluation are $100$, $145$, and $60$ from NYUv2 \cite{silberman2012indoor}, SUNRGB-D \cite{song2015sun} and Hypersim \cite{roberts2021hypersim} datasets. We include both quantitative and qualitative comparisons to evaluate the performance of the proposed volumetric saliency guided summarization method with the methods \cite{lou2020exploiting}, \cite{cong2019going}, \cite{fan2020rethinking}, \cite{zhang2022c}, \cite{zhang2021bilateral}, \cite{li2023mutual}. 

\subsection{Quantitative Performance Comparison}\label{compSOTA}

In this subsection, we present the quantitative performance comparison of the performance of the proposed volumetric saliency guided image summarization method with state-of-the-art methods including \cite{lou2020exploiting}, \cite{cong2019going}, \cite{fan2020rethinking}, \cite{zhang2022c}, \cite{zhang2021bilateral}, \cite{li2023mutual}. 

Since saliency maps guide summarization, an accurate saliency map is required for an effective summary. Hence, we first evaluate the accuracy of the saliency maps generated by the proposed volumetric saliency detection method and the aforementioned methods. The evaluation utilizes metrics `F-Measure', `E-Measure', `S-Measure', and `MAE' to compare the generated saliency maps with ground truth saliency map which are manually annotated. These quantitative comparison metrics are computed using a toolbox provided by \cite{Saliency-Evaluation-Toolbox}.

Table \ref{tab:mos} shows the quantitative comparison between the predicted saliency maps using the methods given in \cite{lou2020exploiting}, \cite{cong2019going}, \cite{fan2020rethinking}, \cite{zhang2022c}, \cite{zhang2021bilateral}, \cite{li2023mutual} and proposed method. 
We notice from the table that the saliency maps generated using color-based saliency method \cite{lou2020exploiting} achieve the lowest value for all comparison metrics for both  NYUv2 \cite{silberman2012indoor} and SUNRGB-D \cite{song2015sun} dataset.

The dependency on color affects the performance for the images containing various illumination conditions. Hence, 
it achieves only $0.477$ value of F-Measure and $0.371$ of MAE for NYUv2 dataset, Whereas $0.461$ and $0.364$ for SUNRGB-D dataset, respectively. Modifying RGB saliency by incorporating the depth information as in work \cite{cong2019going} improved the saliency map in comparison to color-only \cite{lou2020exploiting} but still does not achieve good results in terms of metrics due to incomplete detection of salient objects or the introduction of disturbing backgrounds. Also, the color$+$depth feature-based method \cite{cong2019going} avoids the salient object regions in the background.
Similarly, the recent learning-based methods \cite{zhang2022c} and \cite{li2023mutual} do not perform well for the RGB-D images having salient objects in the background. However, the bilateral attention method presented in \cite{zhang2021bilateral}, predicts the salient objects present in both foreground and background regions. This results in saliency maps that are better than the saliency maps predicted using \cite{lou2020exploiting}, \cite{zhang2022c}, and \cite{li2023mutual}. Due to the direct use of depth image data, this method fails when dealing with coarse depth maps of complex scenarios that provide misleading information.
The deep depth-depurator network \cite{fan2020rethinking} achieves the highest performance among all the other methods. But, the predicted saliency maps contain the background region along with salient regions. 
In contrast, the proposed method shows a performance improvement of $0.189$, $0.159$, $0.208$, and $0.105$ in  F-Measure, E-Measure, S-Measure, and MAE for NYUv2 \cite{silberman2012indoor} dataset, while marginal improvements in case of SUNRGB-D \cite{song2015sun} dataset. 
This improvement is achieved due to the inclusion of a volumetric saliency map that explores the scene-defining salient objects in both foreground and background regions. And the use of a probabilistic framework to delineate the boundary of the salient objects makes the proposed method suitable in the case of coarse depth maps in complex scenes.
   
\begin{table}[!hbt]
\centering
\caption{F-Measure, E-Measure, S-Measure, and MAE comparison of proposed work with SOTA methods on Hypersim \cite{roberts2021hypersim} dataset.}
\label{tab:hyp}
\renewcommand{\arraystretch}{1.3}
\resizebox{\columnwidth}{!}{%
\begin{tabular}{l|ccccccc}
\hline
F-Measure ($\uparrow$)& $0.148$ & $0.217$  & $0.236$ & $0.138$ & $0.188$ &  $0.215$ & $\textbf{0.720}$ \\
 E-Measure ($\uparrow$)  & $0.599$ & $0.609$ & $0.611$ & $0.544$ & $0.561$ & $0.508$ & $\textbf{0.753}$ \\ 
  S-Measure ($\uparrow$) & $0.393$ & $0.428$ & $0.458$ & $0.409$ & $0.459$ & $0.421$ & $\textbf{0.658}$ \\ 
 MAE {($\downarrow$)} & $0.214$  & $0.176$  & $0.184$ & $0.210$ & $0.164$ & $0.166$ & $\textbf{0.156}$ \\ \hline
\end{tabular}
}
\end{table}

Furthermore, to evaluate the performance of the proposed method on a synthetic dataset containing more diverse scenarios, a challenging dataset called Hypersim \cite{roberts2021hypersim} is considered. Hypersim \cite{roberts2021hypersim} is a photorealistic synthetic dataset consisting of highly cluttered scenes containing various lighting and shading conditions. As evident from the Tab. \ref{tab:hyp}, almost all methods including \cite{lou2020exploiting}, \cite{cong2019going}, \cite{fan2020rethinking}, \cite{zhang2022c}, \cite{zhang2021bilateral}, \cite{li2023mutual} does not perform well. One of the reasons for such poor performance is that the synthetic indoor RGB-D images are not natural and contain severe color artifacts, which misleads the saliency detection. While the proposed method can still effectively detect the salient objects in such scenes. The reason is that the probabilistic framework used in the proposed method effectively segments out the salient objects from other regions in these challenging scenarios.

\begin{figure}[!ht]
  \centering
   \includegraphics[width=0.99\linewidth, height=7.2in]{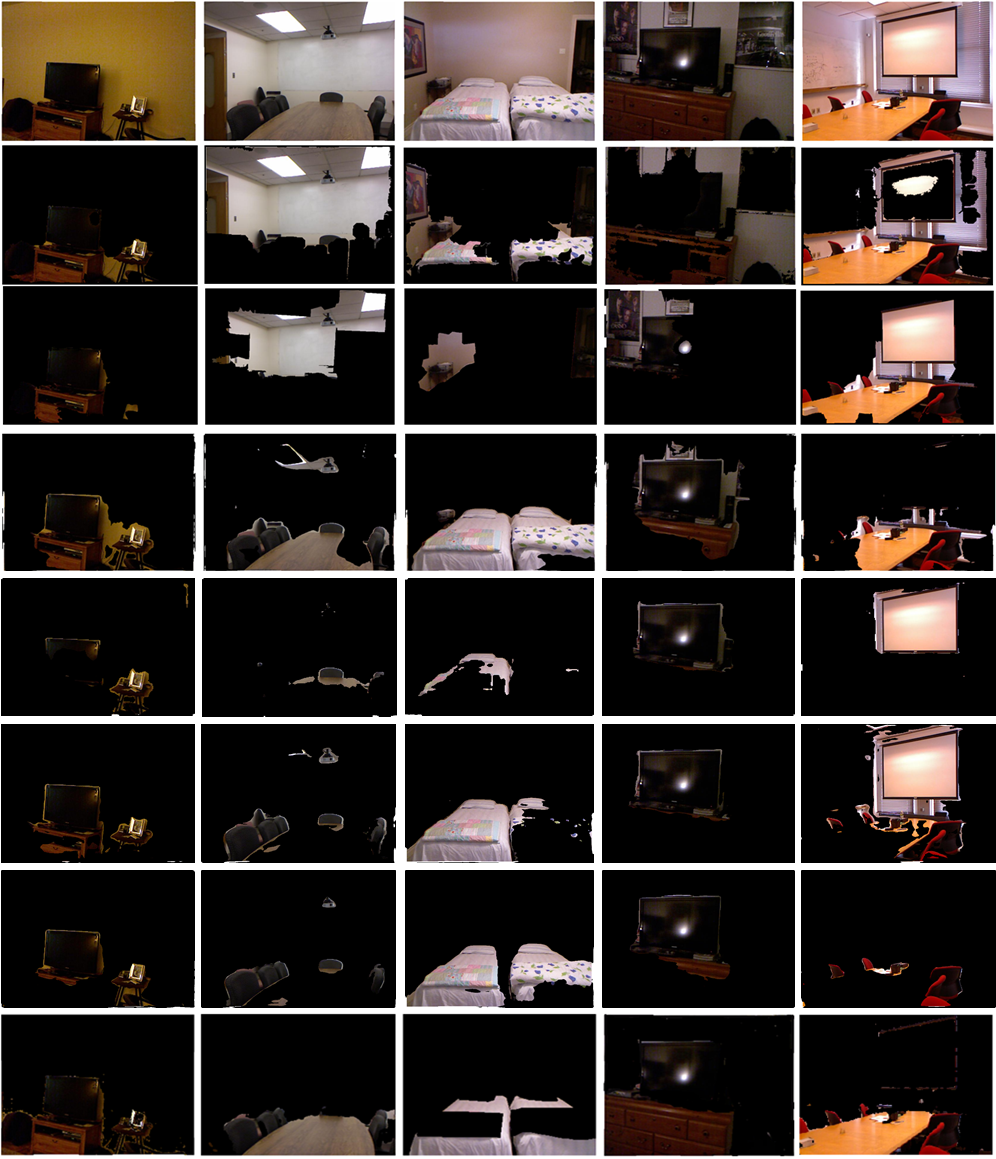}
   \vspace{-0.2cm}
   \caption{Saliency map comparison with SOTA methods. Top row: RGB image. Second to Bottom row: \cite{lou2020exploiting}, \cite{cong2019going}, \cite{fan2020rethinking}, \cite{zhang2022c}, \cite{zhang2021bilateral}, \cite{li2023mutual}, and Proposed method respectively.}
    \label{fig:qua}
\end{figure}

Further, the Mean Opinion Score (MOS) framework is utilized for the performance comparison to evaluate the adequacy of generated summaries from different methods. Methods \cite{lou2020exploiting}, \cite{cong2019going}, \cite{fan2020rethinking}, \cite{zhang2022c}, \cite{zhang2021bilateral}, \cite{li2023mutual} are selected for the comparison of MOS with proposed method. We collected the scores of $38$ viewers on the summaries of $60$ images obtained from all the test methods. Viewers were asked to pick out the objects that they thought were scene-defining in each image. Viewers were subsequently asked to match whether the obtained summaries contained all those scene-defining objects with well-defined boundaries. And then assign a score to each summary on a scale of $1$ to $5$ based on its representativeness of the scene. Later, the average of these scores is computed to get the MOS value between $[0,1]$.


\begin{table}[!hbt]
    \centering
     \caption{Summarization methods performance comparison based on MOS.}
    \label{tab:avgmos}
    \renewcommand{\arraystretch}{1.1}
    \resizebox{\columnwidth}{!}{%
   \begin{tabular}{cccccccc}
    \hline
    Methods &\cite{lou2020exploiting} & \cite{cong2019going} & \cite{fan2020rethinking} & \cite{zhang2022c} & \cite{zhang2021bilateral} & \cite{li2023mutual} & \begin{tabular}[c]{@{}l@{}}Proposed\\ Method\end{tabular} \\  \hline
     MOS & $0.56$ & $0.68$ & $0.80$ & $0.48$ & $0.73$ & $0.81$ & $0.89$\\ \hline
    \end{tabular}
    }
\end{table}

Table \ref{tab:avgmos} presents the comparison of MOS obtained for the summaries generated by methods given in \cite{lou2020exploiting}, \cite{cong2019going}, \cite{fan2020rethinking}, \cite{zhang2022c}, \cite{zhang2021bilateral}, \cite{li2023mutual}, and the proposed image summarization method. We conclude from the table that the proposed volumetric saliency guided image summarization method outperforms others by $0.09$. Thus, we conclude that the proposed volumetric saliency guided results in the adequate and effective summary.


\subsection{Qualitative Performance Comparison}\label{compSOTAqua}

We also present the qualitative comparison between the proposed volumetric saliency guided image summarization method and methods in \cite{lou2020exploiting}, \cite{cong2019going}, \cite{fan2020rethinking}, \cite{zhang2022c}, \cite{zhang2021bilateral}, and \cite{li2023mutual}. Figure \ref{fig:qua} shows the image summaries obtained using methods given in \cite{lou2020exploiting}, \cite{cong2019going}, \cite{fan2020rethinking}, \cite{zhang2022c}, \cite{zhang2021bilateral}, \cite{li2023mutual}, and proposed volumetric saliency guided image summarization method for a few randomly selected RGB-D images. The figure shows that the image summaries obtained using color features \cite{lou2020exploiting} contain the bright regions as salient regions, including walls, and avoid the scene-defining objects (see the second row). Whereas the summaries obtained using color$+$depth feature-based method \cite{cong2019going} consider foreground as salient regions and avoid the background regions.
The image summaries obtained using deep depth-depurator network \cite{fan2020rethinking}, learning-based methods \cite{zhang2022c, li2023mutual}, and bilateral attention method \cite{zhang2021bilateral} contain background interference in salient objects and are mainly focused on foreground regions.
However, the image summaries produced by the proposed image summarization approach include almost all scene-characterizing objects regardless of their position (\ie either present in the foreground or background), unlike others that include regions other than explicitly scene-defining regions. Additionally, the image summaries produced by the proposed method contain complete salient objects with clear object boundaries, and non-salient regions are well suppressed during the saliency detection. The reason is that the probabilistic framework used for segmentation before volumetric saliency guided summary generation in the proposed work effectively delineates the boundary of the salient objects.

\begin{table}[!hbt]
    \centering
    \caption{Sample results for scene classification task using visual summaries.}
     \label{fig:sceneclass}
     \resizebox{\columnwidth}{!}{%
     \begin{tabular}{llllllll}
\hline
\multirow{2}{*}{\begin{tabular}[c]{@{}l@{}}RGB images\\ with labels\end{tabular}} & \multicolumn{6}{c}{Predicted scene labels} \\
 & \cite{lou2020exploiting} & \cite{cong2019going} & \cite{fan2020rethinking} & \cite{zhang2022c} & \cite{zhang2021bilateral} & \cite{li2023mutual} & \begin{tabular}[c]{@{}l@{}}Proposed\end{tabular}\\
\hline
\\
\begin{tabular}[c]{@{}l@{}}
\begin{minipage}[b]{0.3\linewidth}
 \begin{overpic}[width=\textwidth]{376.png}
 \put (2,61) {\fcolorbox{black}{black!15}{\color{blue}{\textbf{home-office}}}}
 \end{overpic}
 \end{minipage}
 \end{tabular} & HO & F & B & B & SD & SD & HO \\

\begin{tabular}[c]{@{}l@{}}
\begin{minipage}[b]{0.3\linewidth}
\begin{overpic}[width=\textwidth]{5.png}
\put (2,61) {\fcolorbox{black}{black!15}{\color{blue}{\textbf{office}}}}
\end{overpic}
\end{minipage}
\end{tabular} & KT & B & B & B & CF & CR & F \\

\begin{tabular}[c]{@{}l@{}}
\begin{minipage}[b]{0.3\linewidth}
\begin{overpic}[width=\textwidth]{977.png}
\put (2,61) {\fcolorbox{black}{black!15}{\color{blue}{\textbf{bedroom}}}}
\end{overpic}
\end{minipage}
\end{tabular} & B &  B & B & B & B & B & B \\

\begin{tabular}[c]{@{}l@{}} \begin{minipage}[b]{0.3\linewidth}
\begin{overpic}[width=\textwidth]{1140.png}
 \put (2,61) {\fcolorbox{black}{black!15}{\color{blue}{\textbf{bedroom}}}}
\end{overpic}
\end{minipage}
\end{tabular} & CR & B & F & F & LR & B & B \\

\begin{tabular}[c]{@{}l@{}} \begin{minipage}[b]{0.3\linewidth}
\begin{overpic}[width=\textwidth]{36.png}
\put (2,61) {\fcolorbox{black}{black!15}{\color{blue}{\textbf{office}}}}
\end{overpic}
\end{minipage} 
\end{tabular} & RR & B & RR & B & LR & LR & LR\\

\begin{tabular}[c]{@{}l@{}} \begin{minipage}[b]{0.3\linewidth}
\begin{overpic}[width=\textwidth]{1311.png}
\put (2,61) {\fcolorbox{black}{black!15}{\color{blue}{\textbf{living room}}}}
\end{overpic}
\end{minipage}  
\end{tabular} & B & B & B & F & LR & LR & LR\\
 \hline
\end{tabular}%
}
B= bedroom, LR=living room, RR=reception room, CF=cafe, SD=study, HO= home-office, CR=classroom, F=office, KT= kitchen. 
\end{table}


\subsection{Performance Comparison for RGB-D Indoor Scene Classification Task}

We also compare the adequacy of the generated summary for tasks such as scene classification. 

For this purpose, we trained a SVM with Bag of visual words (BOVW) \cite{csurka2004visual} features on the $27$ scene categories of the NYUv2 dataset \cite{silberman2012indoor}. The training data consists of the $1449$ original image of the dataset. The trained model is utilized to predict the scene category for the images present in the test sets. 
For test sets creation, a set of images is taken from all three datasets (\ie NYUv2 \cite{silberman2012indoor}, SUNRGB-D \cite{song2015sun} and Hypersim \cite{roberts2021hypersim}). Hence, a set of $200$ original RGB images is formed. Each of these images is then summarized using summarization methods to get a summary of each image in the set. The seven summarization methods including \cite{lou2020exploiting}, \cite{cong2019going}, \cite{fan2020rethinking}, \cite{zhang2022c}, \cite{zhang2021bilateral}, \cite{li2023mutual}, and proposed method are used. Consequently, $7$ test sets are formed that contain the saliency guided image summaries of each image given in the test set. Lastly, these summaries are given to trained SVM to predict the scene category for each image of the obtained test set.

Table \ref{fig:sceneclass} presents a few examples of the indoor scene classification task for a few randomly selected images. The table shows that the proposed volumetric saliency guided image summaries provide a more adequate summary for the scene classification compared to other test methods. Because the image summaries produced by the proposed method contain all scene-defining objects that are helpful in predicting the scene category.

For quantitative comparison of the classification task, accuracy $Acc=\frac{\# \text{correct predictions}}{\# \text{total test images}}$ is utilized. 
\begin{table}[!hbt]
    \centering
     \caption{Comparison of accuracy for scene classification task using visual summaries.}
     \label{tab:SCcomp}
     \renewcommand{\arraystretch}{1.1}
     \resizebox{\columnwidth}{!}{%
     \begin{tabular}{cccccccc}
    \hline
    Methods &\cite{lou2020exploiting} & \cite{cong2019going} & \cite{fan2020rethinking} & \cite{zhang2022c} & \cite{zhang2021bilateral} & \cite{li2023mutual} & \begin{tabular}[c]{@{}l@{}}Proposed\\ Method\end{tabular} \\  \hline
    $Acc$ & $0.57$ & $0.53$ & $0.62$ & $0.55$ & $0.59$ & $0.60$ & $0.82$\\ \hline
    \end{tabular}
    }
\end{table}
Table \ref{tab:SCcomp} shows the quantitative comparison of obtained accuracy for the scene classification task. We notice from the table that the minimum scene classification accuracy is achieved by the test set consisting of image summaries obtained using the method given in \cite{cong2019going}. In contrast, the proposed work achieves a performance improvement of $0.20$. This demonstrates the superior performance of the proposed volumetric saliency guided image summarization method over SOTA methods. Thus, we conclude that the RGB-D indoor scene classification task can be performed efficiently using the unique summary produced by the proposed volumetric saliency guided image summarization method.

\subsection{Further Analysis}
\label{ablation studies}

In order to demonstrate the significance of the proposed novel components in the proposed summarization method, this subsection presents the variation in results due to the presence of these components.

\begin{figure}[!hbt]
    \centering
    \includegraphics[width=0.98\linewidth, height=1.4in]{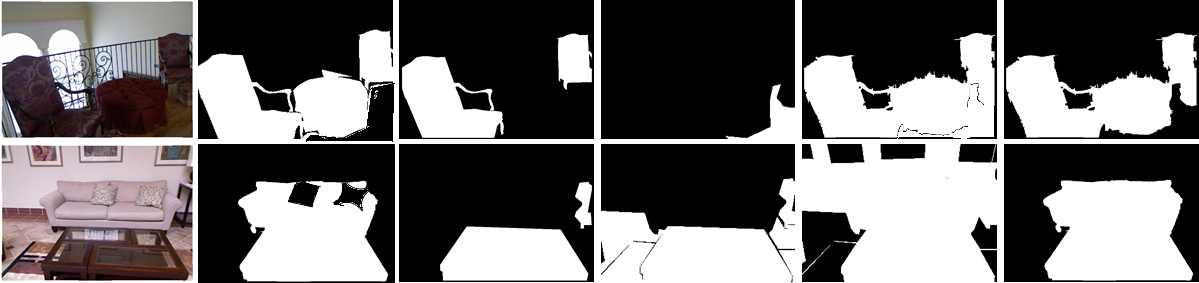}
    \centerline{\footnotesize{(a) \hspace{0.37in} (b) \hspace{0.37in} (c) \hspace{0.37in} (d) \hspace{0.37in} (e) \hspace{0.37in} (f)}}
    \caption{Illustration of impact of using different saliency maps (a) RGB image, (b) GT, (c)-(f) Saliency maps computed using contrast, depth, volume, proposed method.}
    \label{fig:ablation}
\end{figure}

Figure \ref{fig:ablation} shows the predicted saliency maps using contrast, depth, volumetric information, and volumetric information+object classifier.
We observed from the figure that the contrast-based saliency map fails to completely detect all the salient objects. 
Depth-based saliency defines the foreground regions as salient and the scene-defining objects present far (in the background) as non-salient.
Alternatively, the volumetric saliency predicts all the scene-defining objects as salient regardless of their position in the image. Thus, this illustrates the superior performance of volumetric saliency over contrast-based and depth-based saliency. 

Based on volumetric information alone, the saliency detection method detects redundant regions like walls and floors as salient regions. Whereas Saliency detection based on volumetric information and an object classifier $\mathcal{C}$ eliminates these unwanted regions. Hence, the predicted saliency map contains only the scene-defining objects as salient regions and the unwanted regions as non-salient regions. Thus, this indicates the significance of using an object classifier with volumetric information for saliency detection.

\begin{table}[!hbt]
    \centering
    \caption{Comparison of performance metrics for contrast-based saliency, depth-based saliency, volumetric saliency, and volumetric saliency$+$object classifier.}
    \label{tab:volcomp}
    \renewcommand{\arraystretch}{1.3}
    \begin{tabular}{c|cccc}
    \hline
    Metric & $E1$ & $E2$ & $E3$ & $E4$\\ \hline
    F-Measure ($\uparrow$)  & $0.346$ & $0.363$ & $0.631$ & $0.899$\\    
    E-Measure ($\uparrow$)  & $0.259$ & $0.509$ & $0.655$ & $0.892$\\ 
    S-Measure ($\uparrow$)  & $0.168$ & $0.394$ & $0.619$ & $0.878$\\ 
    MAE {($\downarrow$)}  & $0.670$ & $0.367$ & $0.203$ & $0.074$\\ \hline
    \end{tabular}
\end{table}

Table  \ref{tab:volcomp} lists the computed metrics value for predicted saliency maps using contrast-based saliency “$E1$”, depth-based saliency “$E2$”, volumetric saliency “$E3$”, and volumetric saliency$+$object classifier-based “$E4$”. As indicated in the table, volumetric saliency “$E3$” achieves better performance than both “$E1$” and “$E2$” in terms of quantitative measures. While “$E4$” achieves the increment in the value of metrics F-Measure, E-Measure, and S-Measure by $0.268$, $0.237$, and $0.259$, respectively. In contrast, the decrease of $0.129$ in MAE metric. This demonstrates the superior performance of the volumetric saliency$+$object classifier over others.

\section{Conclusion}
\label{Conclusion}
This paper presents a novel method for RGB-D image summarization using a volumetric saliency guide for indoor scene classification. The proposed method utilizes additional spatial and directional features along with the conventional visual features e.g. color and texture, to segment out the objects from the RGB-D images in a probabilistic fusion framework. Further, a volumetric saliency guide is utilized to determine the scene-defining summary using visual saliency. Thus, the proposed two-stage fusion framework of geometric and visual features results in a highly effective summarization framework for indoor RGB-D scenes for scene identification purposes. The qualitative and quantitative analysis verify the efficacy of the proposed method compared to state-of-the-art methods. In the future, we would like to extend the proposed method for the 3D spatial data.

\section{Acknowledgement}

This work has been partially supported under SRG grant with grant no. {\it{SRG/2020/000783}} received from SERB, Govt. of India. 



\bibliographystyle{IEEEtran}
\bibliography{refs}

\begin{IEEEbiography}[]{Preeti Meena} is currently pursuing the Ph.D. degree with the Department of Electrical Engineering, IIT Jodhpur, India. Her research interests include signal processing, image and video processing.
\end{IEEEbiography}

\vspace{11pt}
\begin{IEEEbiography}[]{Himanshu Kumar} (Member, IEEE) received the Ph.D. degree from the Department of Electrical Engineering, IIT Kanpur, India.

He is currently an Assistant Professor with the Department of Electrical Engineering, IIT Jodhpur, India. His research interests include image and video processing, computer vision, optimization, computational imaging, and machine learning.
\end{IEEEbiography}

\vspace{11pt}
\begin{IEEEbiography}[]{Sandeep Kumar Yadav} (Member, IEEE) recieved the Ph.D. degree in Electrical Engineering from Indian Institute of Technology Kanpur (IIT Kanpur), India, in 2010. 

He is currently an Associate Professor with the Department of Electrical Engineering, IIT Jodhpur, India. His research interests include capabilities and limits of Signal Processing, Condition Monitoring, Communication systems, Image Processing, Blind Source Separation, artificial neural network and Healthcare applications.

Dr. Yadav was the recipient of SSI Young Scientist Award (2011), NI GSD Award (2012), IBM shared Research University Award (2014), and IIT Jodhpur teaching excellence award (2020). 
\end{IEEEbiography}

\vfill

\end{document}